\documentclass[letterpaper]{article} 
\usepackage{aaai2026}  
\usepackage{times}  
\usepackage{helvet}  
\usepackage{courier}  
\usepackage[hyphens]{url}  
\usepackage{graphicx} 
\usepackage{natbib}  
\usepackage{caption} 
\usepackage{algorithm}
\usepackage{algorithmic}
\usepackage{times}
\usepackage{soul}
\usepackage{url}
\usepackage{amsmath}
\usepackage{amsthm}
\usepackage{algorithm}
\usepackage{algorithmic}
\usepackage[switch]{lineno}
\usepackage{listings}
\usepackage{adjustbox} 
\usepackage{cleveref} 
\usepackage{multirow} 
\usepackage{array}  
\usepackage{color} 
\usepackage{xcolor}
\usepackage{helvet}
\usepackage[T1]{fontenc} 
\usepackage{bbding} 
\usepackage{booktabs} 
\usepackage{tcolorbox}
\tcbuselibrary{breakable}
\usepackage{caption}
\usepackage{courier}
\usepackage{lipsum}
\usepackage{amsmath} 
\usepackage{graphicx} 
\usepackage{enumitem} 
\usepackage{amsmath}
\usepackage{amsfonts}
\usepackage{amssymb}
\usepackage[table]{xcolor}
\usepackage{enumitem}
\usepackage{amssymb} 
\usepackage{amsmath}
\usepackage{nopageno}
\usepackage{colortbl}

\newlist{checklist}{itemize}{1}
\setlist[checklist,1]{
  label={\(\square\)},
  labelwidth=1em,
  labelsep=1em,
  left=0pt,
  itemsep=0pt
}
\makeatletter
\renewcommand\@makefnmark{}
\makeatother
\usepackage{newfloat}
\usepackage{listings}
\DeclareCaptionStyle{ruled}{labelfont=normalfont,labelsep=colon,strut=off} 
\floatstyle{ruled}
\newfloat{listing}{tb}{lst}{}
\floatname{listing}{Listing}
\title{
Relink: Constructing Query-Driven Evidence Graph On-the-Fly for GraphRAG
}

\author {
    Manzong Huang\textsuperscript{\rm 1},
    Chenyang Bu\textsuperscript{\rm 1*},
    Yi He\textsuperscript{\rm 2},
    Xingrui Zhuo\textsuperscript{\rm 1},
    Xindong Wu \textsuperscript{\rm 1*}
}
\affiliations {
    \textsuperscript{\rm 1}Key Laboratory of Knowledge Engineering with Big Data (the Ministry of Education of China), School of Computer Science and Information Engineering, Hefei University of Technology, China\\
    \textsuperscript{\rm 2}Department of Data Science, College of William and Mary, USA\\
    \{manzonghuang, zxr\}@mail.hfut.edu.cn,
yihe@wm.edu,
\{chenyangbu, xwu\}@hfut.edu.cn
}

\usepackage{bibentry}

\begin{document}

\maketitle

\begin{abstract}

Graph-based Retrieval-Augmented Generation (GraphRAG) mitigates hallucinations in Large Language Models (LLMs) by grounding them in structured knowledge. However, current GraphRAG methods are constrained by a prevailing \textit{build-then-reason} paradigm, which relies on a static, pre-constructed Knowledge Graph (KG). This paradigm faces two critical challenges. First, the KG's inherent incompleteness often breaks reasoning paths. Second, the graph’s low signal-to-noise ratio introduces distractor facts, presenting query-relevant but misleading knowledge that disrupts the reasoning process.
To address these challenges, we argue for a \textit{reason-and-construct} paradigm and propose Relink, a framework that dynamically builds a query-specific evidence graph. To tackle incompleteness, \textbf{Relink} instantiates required facts from a latent relation pool derived from the original text corpus, repairing broken paths on the fly. To handle misleading or distractor facts, Relink employs a unified, query-aware evaluation strategy that jointly considers candidates from both the KG and latent relations, selecting those most useful for answering the query rather than relying on their pre-existence. This empowers Relink to actively discard distractor facts and construct the most faithful and precise evidence path for each query.
Extensive experiments on five Open-Domain Question Answering benchmarks show that Relink achieves significant average improvements of 5.4\% in EM and 5.2\% in F1 over leading GraphRAG baselines, demonstrating the superiority of our proposed framework. 

\end{abstract}

\begin{links}
\link{Code}{https://github.com/DMiC-Lab-HFUT/Relink}
\end{links}

\section{Introduction}
\footnotetext{* Corresponding authors.}

\begin{figure}[t]
 \centering 
 \includegraphics[width=0.47\textwidth]{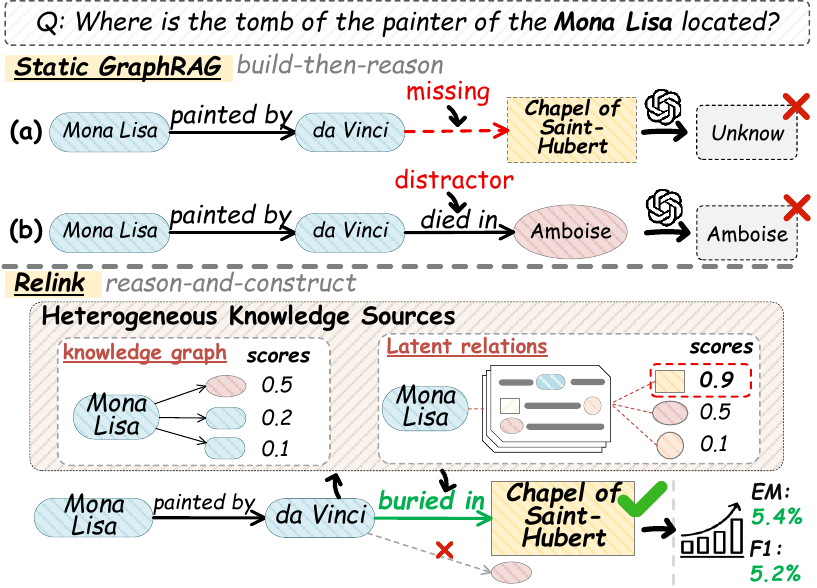}
  \vspace{-0.5em}
 \caption{
Static GraphRAG failures vs. Relink's Dynamic Construction. Pre-built knowledge graphs cause two critical failures in GraphRAG: (a) missing links breaking reasoning paths, and (b) distractor facts (query-relevant but goal-misaligned). In contrast, our \textit{reason-and-construct} approach, Relink, addresses both by discarding distractor facts and dynamically instantiating missing ones from the latent relations derived from the original text corpus.
}
 \label{fig:motivation}
 \vspace{-1em}
\end{figure}

Despite the impressive performance in open-domain question answering (ODQA) \cite{GPT-3,ODQA-review}, large language models (LLMs) are prone to factual errors—known as hallucinations~\cite{roadmap, Hallucination}. Such errors often arise from their over-reliance on internal parametric knowledge.
To mitigate this, Retrieval-Augmented Generation (RAG) grounds LLMs in external knowledge~\cite{RAG-review}. 
GraphRAG~\cite{roadmap,graph_review} further advances this approach by utilizing Knowledge Graph (KG) structures to improve multi-hop query resolution through explicit relational reasoning.

However, this progress has also exposed a fundamental constraint shared by all current GraphRAG methods: the dominant \textbf{\textit{build-then-reason}} paradigm~\cite{TCR-QF,Think-on-Graph}. This paradigm, which relies on a pre-constructed, static KG, faces two critical challenges.

The first challenge is \textit{the inherent incompleteness of KGs}.  
Static KGs inherently suffer from incomplete coverage due to evolving knowledge and extraction errors~\cite{KGCon_review,GoG,Improving_incomplete}.  
To address this, approaches such as Knowledge Graph Completion (KGC)~\cite{GuoWZLX23,LKD-KGC} and LLM-based KG construction~\cite{KGGen,SAC-KG,AutoKG} attempt to densify the graph in advance.  
However, this "global completion" strategy often fails to provide the necessary "local" facts for a given query, causing the reasoning chains to remain fragile.

The second challenge is the \textit{low signal-to-noise ratio}, characterized by an abundance of query-relevant yet distracting facts. 
General-purpose KGs contain numerous facts that may topically relate to queries but lack precise answer utility.
As illustrated in Figure~\ref{fig:motivation}(b), the relation \textit{died in} (versus \textit{buried in}) exemplifies this issue: highly query-relevant yet functionally distracting.
Existing methods, such as retrieval refinement and textual KG supplementation~\cite{ToG2,TCR-QF}, remain fundamentally static KG-dependent. This leaves them vulnerable to error propagation, where misleading facts are amplified during reasoning.

These challenges reveal a fundamental flaw in the prevailing \textbf{\textit{build-then-reason}} paradigm, which relies on a one-graph-fits-all approach. Current methods are constrained by static KGs rather than adaptively serving query-specific user queries. To overcome this, we advocate a paradigm shift to \textbf{\textit{reason-and-construct}}~\cite{Wu2025BEKO,Multimodal_GraphRAG}, dynamically constructing compact and query-aligned evidence graph that ensures precise reasoning path alignment.

To realize this new {\textit{reason-and-construct}} paradigm, we propose \textbf{Relink}, a framework designed to address both aforementioned challenges via complementary mechanisms: (1) For KG incompleteness, Relink dynamically instantiates missing relations from the latent relations derived from the original text corpus. A high-precision KG serves as a skeletal backbone, providing a reliable foundation that inherently minimizes the presence of distractor facts. To complement its limited coverage, a high-recall latent relation pool built from entity co-occurrences in the text corpus supplies additional candidate links. This enables Relink to dynamically repair broken paths by constructing the missing facts required to answer a query.
(2) For distractor noise, Relink adopts a unified evaluation strategy. At each step, a query-aware ranker assesses a pool of competing candidates, drawn from both existing KG facts and potential relations. The ranker's choice is based on a candidate's utility for answering the query, not its pre-existence. This enables Relink to actively discard misleading paths (the \texttt{"died in"} relation in \Cref{fig:motivation}(b)) and instead construct the most relevant ones (the \texttt{"buried in"} relation in \Cref{fig:motivation}), ensuring that the evidence graph remains precise and free of noise from the outset.

Extensive experiments on five ODQA benchmarks show that Relink outperforms leading GraphRAG baselines, achieving average gains of 5.4\% in EM and 5.2\% in F1. This provides compelling evidence for the superiority of the proposed paradigm.

\smallskip\noindent
{\bf The specific contributions} of this paper are as follows:
\begin{itemize}
\item[1)] We systematically analyze the prevailing \textit{build-then-reason} paradigm, identifying its core failures in handling \textit{KG incompleteness}, and more critically in navigating the distractor facts that stem from \textit{low signal-to-noise ratios}.
\item[2)] Following the idea of reason-and-construct, we propose the framework Relink to dynamically construct evidence graphs through unified evaluation of explicit and latent knowledge, achieving on-the-fly path repair and distractor filtering.
\item[3)] Extensive experiments on five ODQA benchmarks show that Relink outperforms leading GraphRAG baselines by an average of 5.4\% in EM and 5.2\% in F1.
\end{itemize}

\section{Related Work}

GraphRAG enhances LLM reasoning by grounding it in KGs~\cite{roadmap,graph_review_2,DBLP:conf/www/ZhuoWWP025}, proving particularly effective for complex, multi-hop question answering~\cite{In-depth}. However, the prevailing methods are built upon a \textit{build-then-reason} paradigm, where reasoning is performed over a pre-constructed, static KG. This approach faces two challenges: the KG’s inherent incompleteness, which breaks potential reasoning paths, and the presence of misleading distractor facts, which disrupt the reasoning process~\cite{incomplete_of_kg,GiVE,IncompleteKG,GoG}.

Existing methods attempt to address these issues within the constraints of this paradigm. One line of work focuses on alleviating KG incompleteness ahead of time through "global completion", either via KGC techniques~\cite{GuoWZLX23,LKD-KGC,KGCon_review,DBLP:journals/tkde/ZhuoWZW24} or LLM-based KG construction~\cite{KGGen,SAC-KG,AutoKG}. However, these query-agnostic strategies densify the graph indiscriminately and often fail to satisfy the "local" facts needed for the specific query, leaving reasoning paths fragile when key links are missing.
Another line of work aims to improve information relevance by refining retrieval~\cite{LightRAG,G-Retriever,GraphRAG,StructGPT,KG-Retriever,DBLP:conf/ijcai/ZhuoPWW0LW025} or supplementing evidence graphs with additional text~\cite{TCR-QF,HOLMES,GraphRAG}. While these approaches enhance retrieval or ranking to reduce noise, they remain fundamentally tied to the initial graph. Consequently, they struggle to establish new reasoning paths when required links are absent and are still vulnerable to being misled by distractor facts.

In contrast, Relink embodies the \textit{reason-and-construct} paradigm, eschewing reliance on a static graph by dynamically constructing a compact, query-specific evidence graph. Rather than simply traversing a pre-built structure, Relink adopts a unified evaluation strategy at each step, assessing candidates from both the KG and a latent relation pool derived from co-occurrence patterns in the corpus. This enables Relink to instantiate essential links while actively discarding distractors. By doing so, it addresses both KG incompleteness and distractors in KG, ensuring that the resulting reasoning path is both robust and highly relevant.

\section{Proposed Framework}

\begin{figure*}[htbp]
\centering
\includegraphics[width=0.95\textwidth]{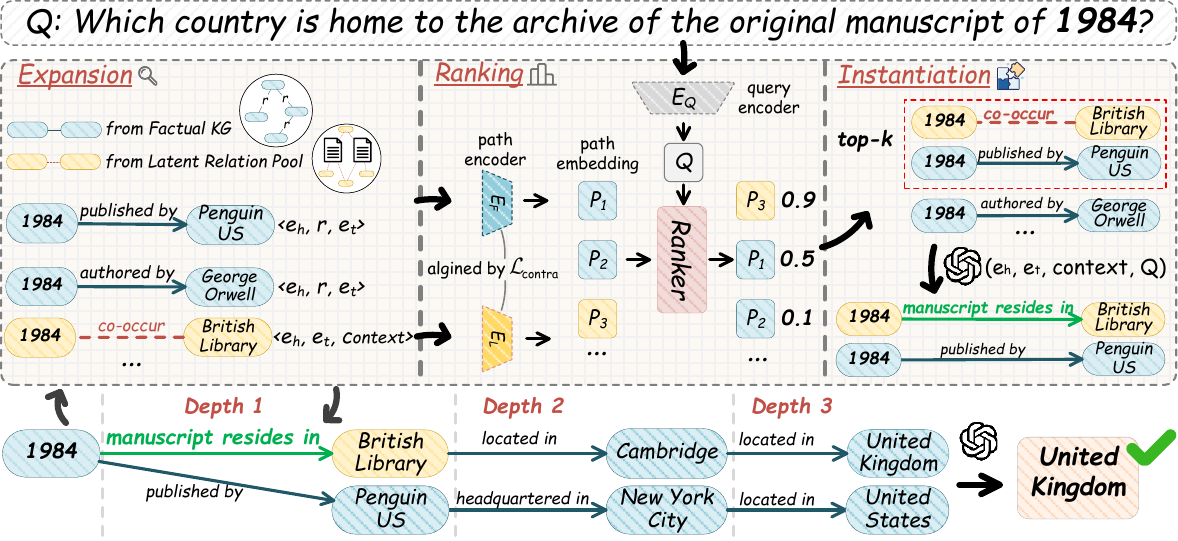}
\caption{Relink's dynamic evidence graph construction. Relink iteratively builds reasoning paths by leveraging candidates from both the explicit KG ($\mathcal{G}_b$), and the latent co-occurrence relation pool ($\mathcal{R}_c$) derived from the corpus. Encoders $E_L$ and $E_F$ project these candidates into a unified semantic space where a query-driven ranker evaluates their relevance. When  latent relations are selected, an LLM instantiates them into factual relations (e.g., \texttt{"manuscript resides in"}) using source context, dynamically repairing missing path segments during construction.}
\label{fig:framework}
\vspace{-1.em}
\end{figure*}

We propose Relink, a framework that embodies the \textit{reason-and-construct} paradigm, with its overall architecture illustrated in Figure~\ref{fig:framework}. Rather than reasoning over a static KG, Relink dynamically constructs a compact, query-specific evidence graph. This approach addresses the challenges of KG incompleteness and distractor facts through two core designs: a Heterogeneous Knowledge Source that integrates complementary sources of candidate facts, leveraging their combined coverage to mitigate incompleteness, and a Query-Driven Dynamic Path Exploration module that selectively navigates these sources to filter out distractors.


\subsection{Heterogeneous Knowledge Source Construction}
To address \textbf{KG incompleteness}, Relink's ability to repair broken paths stems from its heterogeneous knowledge source. This source is designed to balance the precision of a factual KG with the high recall of latent textual relations.

\vspace{0.3em}
\noindent\textbf{High-Precision Factual KG ($\mathcal{G}_b$):} This is a standard KG, $\mathcal{G}_b = (\mathcal{E}, \mathcal{R}_b)$, where $\mathcal{E}$ is the set of entities and $\mathcal{R}_b$ is a set of high-confidence factual relations. This graph is constructed using an LLM-based extractor from the text corpus. As a result, it serves as a reliable yet inherently incomplete backbone for reasoning.

\vspace{0.3em}
\noindent\textbf{High-Recall Latent Relation Pool ($\mathcal{R}_c$):} To address the limitations of incompleteness in $\mathcal{G}_b$, we introduce a high-recall pool of potential relations, $\mathcal{R}_c$, derived from textual entity co-occurrences. This pool serves as the raw material for path repair. Specifically, we identify co-occurring entity pairs $(e_i, e_j)$ within the corpus and filter for meaningful associations using a Pointwise Mutual Information (PMI) threshold:
\[
\text{PMI}(e_i, e_j) = \log \frac{p(e_i, e_j)}{p(e_i)p(e_j)} > \tau_c
\]
PMI is chosen for its ability to capture non-linear co-occurrence patterns while reducing bias toward high-frequency entities. For each valid pair, we encode its context sentence $c_{ij}$ using a pretrained language model, $\text{Encoder}_L$, to generate a dense representation for the latent relation, $\mathbf{r}_{ij} \in \mathbb{R}^d$. Following standard practice~\cite{matchPrompt,PromptORE}, we use the last hidden state of the \texttt{[MASK]} token from a formatted input $s_{ij} = \texttt{[CLS]}~c_{ij}~\texttt{[SEP]}~e_i~\texttt{[MASK]}~e_j~\texttt{[SEP]}$:
\[
\mathbf{r}_{ij} = \text{Encoder}_L(s_{ij})_{[\text{MASK}]}
\]
This collection, $\mathcal{R}_c$, serves as a repository of candidate connections that can be leveraged to dynamically repair and enrich reasoning paths.

\subsection{Query-Driven Dynamic Path Exploration}

This module lies at the heart of Relink, designed to efficiently navigate the search space and tackle the challenge posed by \textbf{distractor facts}. Given a query $q$, it iteratively constructs a set of evidence paths, $\mathcal{P}^*$, by intelligently selecting steps from the heterogeneous knowledge source.

\paragraph{Unified Semantic Space.}
A crucial initial step is to jointly reason over explicit triples from $\mathcal{G}_b$ and latent relations from $\mathcal{R}_c$ by projecting both into a unified semantic space $\mathbb{R}^d$. This unification is essential, as it enables a single ranker to evaluate the facts from diverse sources on a common basis.
\begin{itemize}
    \item[] An \textbf{explicit factual triple} $(h, p, t) \in \mathcal{G}_b$ is linearized into a sequence $s = \texttt{[CLS]}\, h\, p\, t\, \texttt{[SEP]}$, then encoded by $\text{Encoder}_F$. The representation of the $\mathtt{[CLS]}$ token represents the triple: $\mathbf{v}_f = \text{Encoder}_F(s)_{[\mathtt{CLS}]}$.
    \item[] A \textbf{latent relation}'s pre-computed representation $\mathbf{r}_{ij}$ is used directly as its vector $\mathbf{v}_l$.
\end{itemize}
This unification yields a common representation $\mathbf{v}_{\text{edge}}$ for each candidate edge, enabling seamless joint ranking. The semantic coherence of this unified space is maintained during training through a contrastive alignment loss, as described in a subsequent section.

\paragraph{Iterative Path Expansion and Ranking.} 
We employ a beam search algorithm, initiating from the topic entities present in the query $q$ to systematically explore the reasoning space. At each step, we prioritize not only "relevant" but also truly "precise" knowledge from the heterogeneous source, focusing on facts that directly contribute to answering the question while effectively filtering out distractors. The search unfolds in three stages:

\begin{itemize}
\item[] \textbf{Candidate Expansion:} For each partial path $P_{k-1}$ in the beam, we expand a set of candidate extensions by identifying all one-hop neighbors of its last entity. These neighbors are drawn from \textit{both} the explicit graph $\mathcal{G}_b$ and the latent pool $\mathcal{R}_c$.

\item[] \textbf{Query-Driven Ranking:} To efficiently identify the most promising extensions, we employ a \textbf{coarse-to-fine ranking strategy}:\\
    \textbf{Coarse Ranking:} A lightweight, trainable Ranker model first scores all candidate extensions. This ranker is optimized to predict an extension's relevance to the query $q$. This step rapidly filters out a large number of irrelevant or noisy paths, retaining a small set of high-potential candidates.\\
    \textbf{Fine-grained Re-ranking:} The top candidates from the coarse stage are then re-evaluated by an LLM. Using a structured prompt, the LLM assesses the semantic contribution of the edge toward answering the query and provides a relevance score increment $\Delta S$.\\
The final average score for a path \(P_k\) is updated recursively:
\[
\bar{S}(P_k|q) = \frac{(k-1) \cdot \bar{S}(P_{k-1}|q) + \Delta S(e_{\text{new}}\,|\,P_{k-1}, q)}{k}
\]
At each step, the top-\(K\) paths with the highest average scores are retained for the next iteration.

\item[] \textbf{Dynamic Instantiation:} This step serves a dual purpose. Primarily, it repairs paths disrupted by KG incompleteness. At the same time, it acts as a final, precise filter against distractor facts. When a top-ranked path requires the instantiation of a latent relation \(\mathbf{r}_{ij}\), we prompt the LLM with both the source context \(c_{ij}\) and the original query \(q\). This query-aware generation encourages the LLM to produce a factual triple \((h, p, t)\) that is closely aligned with the user's intent, rather than generating a more generic or potentially misleading relation that may also appear in the source text.
\[
(h, p, t) = \text{LLM}_{\text{instantiate}}(e_i, e_j, c_{ij}, q)
\]
This ensures that the newly constructed link is not only a repair, but also the most relevant possible, effectively filtering out less pertinent alternatives.
\end{itemize}
The search process terminates when either a maximum path length is reached or the LLM deems a path complete.

\subsection{Evidence-Grounded Answer Generation}
The output of the exploration phase is a compact, query-specific evidence graph. Each triple in this graph is linked to a single source sentence, either as provenance from \(\mathcal{G}_b\) or as the specific sentence \(c_{ij}\) used for instantiated facts. This graph, along with all associated source sentences, is then provided to a generator LLM. The LLM is prompted with both the original query \(q\) and the structured evidence. By grounding the generation process in the logical reasoning structure and the precise sentence-level sources for each triple, the model produces answers that are not only accurate but also faithful and verifiable.

\subsection{Joint Training Objective}
The framework's learnable components, including the Ranker and the encoders ($\text{Encoder}_F$, $\text{Encoder}_L$), are optimized jointly via a multi-task objective designed to foster both ranking precision and semantic alignment.

\paragraph{Ranking Loss (\(\mathcal{L}_{\text{rank}}\)).} The Ranker is trained to distinguish beneficial reasoning steps from unhelpful ones using a pairwise ranking loss. We construct a dataset of preference tuples \((q, P^+, P^-)\), where path \(P^+\) represents a more direct step toward the correct answer for query \(q\) compared to path \(P^-\). The Ranker is then optimized to assign a higher score to \(P^+\) than to \(P^-\):
$$
\mathcal{L}_{\text{rank}} = \mathbb{E}_{(q, P^+, P^-)} \left[ \max(0, m - S(P^+|q) + S(P^-|q)) \right]
$$
where $m$ is a margin hyperparameter. This loss directly optimizes the model's ability to navigate the search space efficiently.

\paragraph{Contrastive Alignment Loss ($\mathcal{L}_{\text{contra}}$).}
\label{sec:training_objective}
To ensure coherence within the unified semantic space, we align the representations of explicit facts (\(\mathbf{v}_f\)) and their corresponding latent relations (\(\mathbf{v}_l\)) using a contrastive objective. This alignment is crucial for enabling the Ranker to make meaningful comparisons. Specifically, we employ the InfoNCE loss to draw the embeddings of a factual triple and its corresponding latent relation closer together, while simultaneously pushing them apart from \(N\) in-batch negative samples:
$$
\label{eq:contra_loss_resize}
\resizebox{1.0\linewidth}{!}{$
\mathcal{L}_{\text{contra}} = -\mathbb{E} \left[ \log \frac{\exp(s(\mathbf{v}_f, \mathbf{v}_l)/\tau)}{\exp(s(\mathbf{v}_f, \mathbf{v}_l)/\tau) + \sum_{j=1}^N \exp(s(\mathbf{v}_f, \mathbf{v}_j^{-})/\tau)} \right]
$}
$$

where $s(\cdot, \cdot)$ is cosine similarity, $\mathbf{v}_j^{-}$ are negative samples, and $\tau$ is a temperature parameter.

\paragraph{Staged Optimization Strategy.}
We adopt a staged training procedure to stably optimize the two objectives, decoupling the challenges of learning semantic representations and ranking logic. Training alternates between two stages:
\begin{itemize}
    \item \textbf{Ranker Training:} The encoders are frozen while the Ranker is trained for one epoch using \(\mathcal{L}_{\text{rank}}\).
    \item \textbf{Encoder Alignment:} The Ranker is frozen while the encoders (\(\text{Encoder}_F, \text{Encoder}_L\)) are trained for one epoch using \(\mathcal{L}_{\text{contra}}\).
\end{itemize}
This cycle repeats until convergence on the validation set, enabling each component to learn its respective task without interference.


\section{Experiments}

\begin{table*}[!htbp]
\centering
\small
\begin{tabular}{cccccccccccc}
\toprule
\multirow{2}{*}{\textbf{Method Type}} & \multirow{2}{*}{\textbf{Method}} 
& \multicolumn{2}{c}{\textbf{2WikiMultiHopQA}} 
& \multicolumn{2}{c}{\textbf{HotpotQA}}
& \multicolumn{2}{c}{\textbf{ConcurrentQA}}  
& \multicolumn{2}{c}{\textbf{MuSiQue-Ans}} 
& \multicolumn{2}{c}{\textbf{MuSiQue-Full}} \\
\cmidrule(lr){3-4} \cmidrule(lr){5-6} \cmidrule(lr){7-8} \cmidrule(lr){9-10} \cmidrule(lr){11-12}
& & \textbf{EM} & \textbf{F1} & \textbf{EM} & \textbf{F1} & \textbf{EM} & \textbf{F1} & \textbf{EM} & \textbf{F1} & \textbf{EM} & \textbf{F1} \\
\midrule
\multirow{2}{*}{\textbf{LLM only}} 

& \textbf{Deepseek-v3} 
& 0.312 & 0.365
& 0.294 & 0.402 
& 0.114 & 0.170
& 0.066 & 0.167
& 0.078 & 0.164 
 \\

\cmidrule(lr){2-12}
& \textbf{GPT-4o} 
& 0.292 & 0.358
& 0.330 & 0.424 
& 0.086 & 0.158
& 0.106 & 0.213
& 0.106 & 0.214 
 \\

\midrule
\multirow{2}{*}{\textbf{Text-based}} 
& \textbf{Vanilla RAG} 
& 0.264 & 0.295 
& 0.376 & 0.494 
& 0.31 &0.389 
& 0.118 & 0.178
& 0.08 & 0.145 
  \\

\cmidrule(lr){2-12}
& \textbf{RAPTOR}
& 0.467 & 0.548 
& 0.472 & 0.631
& 0.400 & 0.476
& \underline{0.258} & 0.371
& 0.188 & 0.286 
\\
\midrule
\multirow{2}{*}{\textbf{Graph-based}} 
& \textbf{TOG} 
& 0.328 & 0.373 
& 0.241 & 0.292 
& 0.166 & 0.185 
& 0.095 & 0.127
& 0.053 & 0.070 
\\

\cmidrule(lr){2-12}
& \textbf{G-Retriever} 
& 0.362 & 0.455 
& 0.369 & 0.493
& 0.162 & 0.209 
& 0.166 & 0.271
& 0.092 & 0.169 
 \\

\midrule
\multirow{4}{*}{\textbf{Hybrid}} 

& \textbf{LightRAG} 
& 0.300 & 0.364 
& 0.384 & 0.498
& 0.308 & 0.389
& 0.124 & 0.183
& 0.104 & 0.171 
\\
\cmidrule(lr){2-12}
& \textbf{GraphRAG} 
& 0.318 & 0.379 
& 0.45 & 0.569
& 0.398 & 0.475
& 0.203 & 0.358
& 0.138 & 0.206 
\\

\cmidrule(lr){2-12}
 & \textbf{HippoRAG} 
 & \underline{0.578} & \underline{0.684} 
 & \underline{0.498} & \underline{0.647}
 & \underline{0.458} & \underline{0.536}
  & 0.254 & \underline{0.381}
 & \underline{0.190} & \underline{0.298}
 \\

\midrule
\rowcolor{gray!20}
\textbf{Proposed} & \textbf{Relink} 
& \textbf{0.628} & \textbf{0.722} 
& \textbf{0.558} & \textbf{0.704}
& \textbf{0.505} & \textbf{0.596} 
& \textbf{0.304} & \textbf{0.413}
& \textbf{0.252} & \textbf{0.370} 
 \\
\bottomrule
\end{tabular}
\caption{Main performance comparison. Relink consistently outperforms all baseline methods across all datasets, demonstrating the effectiveness of its dynamic, query-driven path repair mechanism. Best results are in \textbf{bold}; second-best are \underline{underlined}.}
\label{tab:main_result}
\vspace{-1em}
\end{table*}

To validate the effectiveness of the proposed \textbf{Relink} framework, we performed a comprehensive evaluation designed to assess its multi-hop reasoning capabilities against leading GraphRAG baseline methods.

\subsection{Experimental Setup}

\textbf{Datasets and Evaluation Metrics.}
We evaluate on five standard multi-hop QA benchmarks: 2WikiMultiHopQA~\cite{2WikiMultiHopQA}, HotpotQA~\cite{HotpotQA}, ConcurrentQA~\cite{ConcurrentQA}, MuSiQue-Ans, and MuSiQue-Full~\cite{musique}. Performance is measured by EM and F1, consistent with prior work~\cite{HOLMES,TCR-QF}. Since LLMs often generate answers accompanied by explanations, all model outputs undergo a uniform post-processing step in which an LLM extracts the standardized answer string before scoring.
\\ \noindent\textbf{Baseline Methods.}
Relink was evaluated against a diverse set of representative baselines spanning the major paradigms in multi-hop QA. For consistency, all graph-based methods were implemented using the unified framework~\cite{In-depth}. The baselines are grouped as follows:
\textbf{(1) LLM Baselines:} Methods using only LLMs, including \texttt{deepseek-v3-0324} and \texttt{gpt-4o-2024-07-06}.
\textbf{(2) Text-based RAG:} Approaches that retrieve chunks from corpora. This includes \textbf{Vanilla RAG} (via LangChain) and \textbf{RAPTOR}~\cite{RAPTOR}, which builds a tree-structured index.
\textbf{(3) Graph-based RAG:} Methods reasoning over pre-constructed KGs, including \textbf{ToG}~\cite{Think-on-Graph} and \textbf{G-Retriever}~\cite{G-Retriever}.
\textbf{(4) Hybrid RAG:} Leading methods combining graph and text retrieval, represented by \textbf{GraphRAG}~\cite{GraphRAG}, \textbf{LightRAG}~\cite{LightRAG} and \textbf{HippoRAG}~\cite{HippoRAG}.
\\ \noindent\textbf{Implementation Details.}
All RAG variants, including Relink, employ \texttt{deepseek-v3-0324} as the backbone LLM to ensure comparability. Following prior works~\cite{HOLMES,TCR-QF}, each method is evaluated on 500 randomly sampled questions from the test split of every dataset to reduce computational costs. 
\subsection{Main Performance Comparison}
Table~\ref{tab:main_result} presents the comparative results, which allow us to address the following Research Question (RQ):
\begin{description}
    \item[RQ1] \textbf{\textit{How does Relink, which follows the "reason-and-construct" paradigm, perform compared to leading GraphRAG methods on ODQA benchmarks?}}
\end{description}
The experimental results offer empirical evidence for the proposed method. Relink consistently outperforms all baselines on five widely used ODQA benchmarks, demonstrating its effectiveness in addressing the incompleteness of knowledge graphs and the presence of distractor facts.

\textbf{Compared to the LLM-only and Text-based RAG baselines}, Relink delivers substantial improvements. For example, on 2WikiMultiHopQA, it achieves an EM score of $0.628$, representing a 115.1\% relative improvement over \texttt{GPT-4o} (0.292).
 It also outperforms the strong \texttt{RAPTOR} baseline, with a relative EM gain of 18.2\% on HotpotQA ($0.558$ vs. $0.472$) and 34.5\% on 2WikiMultiHopQA ($0.628$ vs. $0.467$).
Experimental results show that relying solely on parameterized knowledge or unstructured text falls short for multi-hop reasoning. Multi-hop QA requires not just facts, but also clear relationships and reasoning chains between them. Relink addresses this by constructing structured evidence graphs to explicitly organize information and relationships, significantly improving the accuracy and traceability of complex reasoning.

\textbf{When compared to Graph-based and Hybrid methods}, Relink demonstrates clear advantages. While existing GraphRAG approaches rely on static knowledge graphs, they are often limited by the incompleteness of pre-constructed graphs and the presence of distracting facts. Relink, in contrast, adopts a dynamic \textbf{\textit{reason-and-construct}} paradigm, which enables it to construct query-specific evidence graphs on the fly. 
Empirical results confirm the effectiveness of this approach. For instance, Relink surpasses all graph-based and hybrid baselines, including the leading \texttt{HippoRAG}. On HotpotQA, Relink achieves a 12.0\% relative EM improvement over \texttt{HippoRAG}. The advantage is even more pronounced on challenging datasets such as MuSiQue-Full, where Relink attains a 32.6\% higher EM score ($0.252$ vs. $0.190$). These gains highlight that dynamically constructing evidence graphs enables Relink to more effectively select relevant facts, repair incomplete reasoning chains, and filter out misleading information. 

Overall, these results validate our core hypothesis that dynamic, query-aware graph construction is a more robust and effective strategy for complex multi-hop reasoning than static graph-based methods.

\subsection{Ablation Study}
We conducted a comprehensive ablation study to deconstruct the architecture of \textbf{Relink}. The results presented in Table~\ref{tab:ablation} provide an empirical answer to RQ2.

\begin{description}
    \item[RQ2] \textit{\textbf{How effectively do Relink's core components address the challenges of KG incompleteness and distractor facts?}}
    \vspace{-0.5em}
\end{description}

\begin{table}[h]
\centering
\scriptsize
\begin{tabular}{c@{\hskip 12pt}c@{\hskip 8pt}c@{\hskip 12pt}c@{\hskip 8pt}c}
\toprule
\multirow{2}{*}{\textbf{Method}} & \multicolumn{2}{c}{\textbf{2WikiMultiHopQA}} & \multicolumn{2}{c}{\textbf{HotpotQA}} \\
\cmidrule(lr){2-3} \cmidrule(lr){4-5}
& \textbf{EM} & \textbf{F1} & \textbf{EM} & \textbf{F1} \\
\midrule
\textbf{Relink (Full Model)}
& \textbf{0.628} & \textbf{0.722}
& \textbf{0.558} & \textbf{0.704} \\
\midrule
\quad w/o Explicit Graph ($\mathcal{G}_b$)
& 0.582 & 0.672
& 0.486 & 0.636 \\
\quad w/o Dynamic Repair ($\mathcal{R}_c$)
& 0.616 & 0.714
& 0.526 & 0.680 \\
\quad w/o Query-Driven Ranker
& 0.552 & 0.649
& 0.450 & 0.600 \\

\quad w/o $\mathcal{L}_{\text{contra}}$
& 0.603 & 0.695
& 0.518 & 0.675 \\
\bottomrule
\end{tabular}
\caption{Ablation study of Relink on the 2WikiMultiHopQA and HotpotQA datasets. The removal of any component leads to a notable performance drop, highlighting their individual contributions and synergistic importance.}
\label{tab:ablation}
\end{table}

\noindent\textbf{Heterogeneous Knowledge for Completeness.} Our dual-source approach is designed to combat KG incompleteness. The results confirm its necessity. Removing the latent relation pool ($\mathcal{R}_c$) causes a 5.7\% relative drop in EM on HotpotQA, while removing the explicit graph backbone ($\mathcal{G}_b$) leads to a more severe \textbf{12.9\%} drop. This demonstrates that a reliable factual backbone is essential to ground the reasoning process, while the latent pool is critical for dynamically repairing incomplete reasoning paths. Neither source is sufficient on its own, and their synergy is vital for robust performance.

\noindent\textbf{Query-Driven Ranker for Precision.} The query-driven ranker is our primary mechanism for tackling the low signal-to-noise ratio and filtering distractor facts. We replaced the ranker component with a general method that computes cosine similarity over embeddings from OpenAI’s \texttt{text-embedding-3-small} model to validate its contribution. This replacement results in the most significant performance degradation, causing a \textbf{19.4\%} relative EM drop on HotpotQA. This result highlights that generic semantic similarity captures topical relevance but fails to identify facts that are truly useful for reasoning. In contrast, our ranker is explicitly trained to select evidence that supports the specific reasoning needs of each query. This ability to distinguish "useful" from merely "related" facts is essential for precise, goal-directed evidence graph construction.

\noindent\textbf{Unified Alignment for Synergy.} A unified semantic space is essential for reasoning over heterogeneous knowledge sources. Removing the contrastive alignment loss ($\mathcal{L}_{\text{contra}}$) leads to a 7.2\% relative EM drop on HotpotQA, underscoring the challenge of integrating structured and unstructured evidence. Without alignment, the ranker cannot meaningfully compare facts from different sources, reducing its effectiveness. By aligning representations, the model enables the ranker to evaluate all evidence on a common basis, regardless of origin. This unified alignment is thus crucial for fully leveraging heterogeneous knowledge and achieving effective multi-hop reasoning.

The above ablation studies demonstrate that Relink’s components work in concert to balance reasoning coverage and precision. The latent pool enhances completeness by bridging knowledge gaps. The query-driven ranker delivers precision by filtering this noise and selecting relevant facts, but it relies on a diverse candidate set. The explicit graph ensures reliability by grounding reasoning in verified knowledge. This interplay among components is key to the effectiveness of our \textit{reason-and-construct} approach.

\subsection{Robustness under Knowledge Sparsity}
We assess Relink's performance under increasingly incomplete knowledge to address RQ3:
\begin{description}
\item[RQ3] \textbf{\textit{How does the resilience of the "reason-and-construct" paradigm to knowledge sparsity compare against the inherent brittleness of static reasoning approaches?}}
\end{description}

\begin{figure}[ht]
\centering
\includegraphics[width=1.0\linewidth]{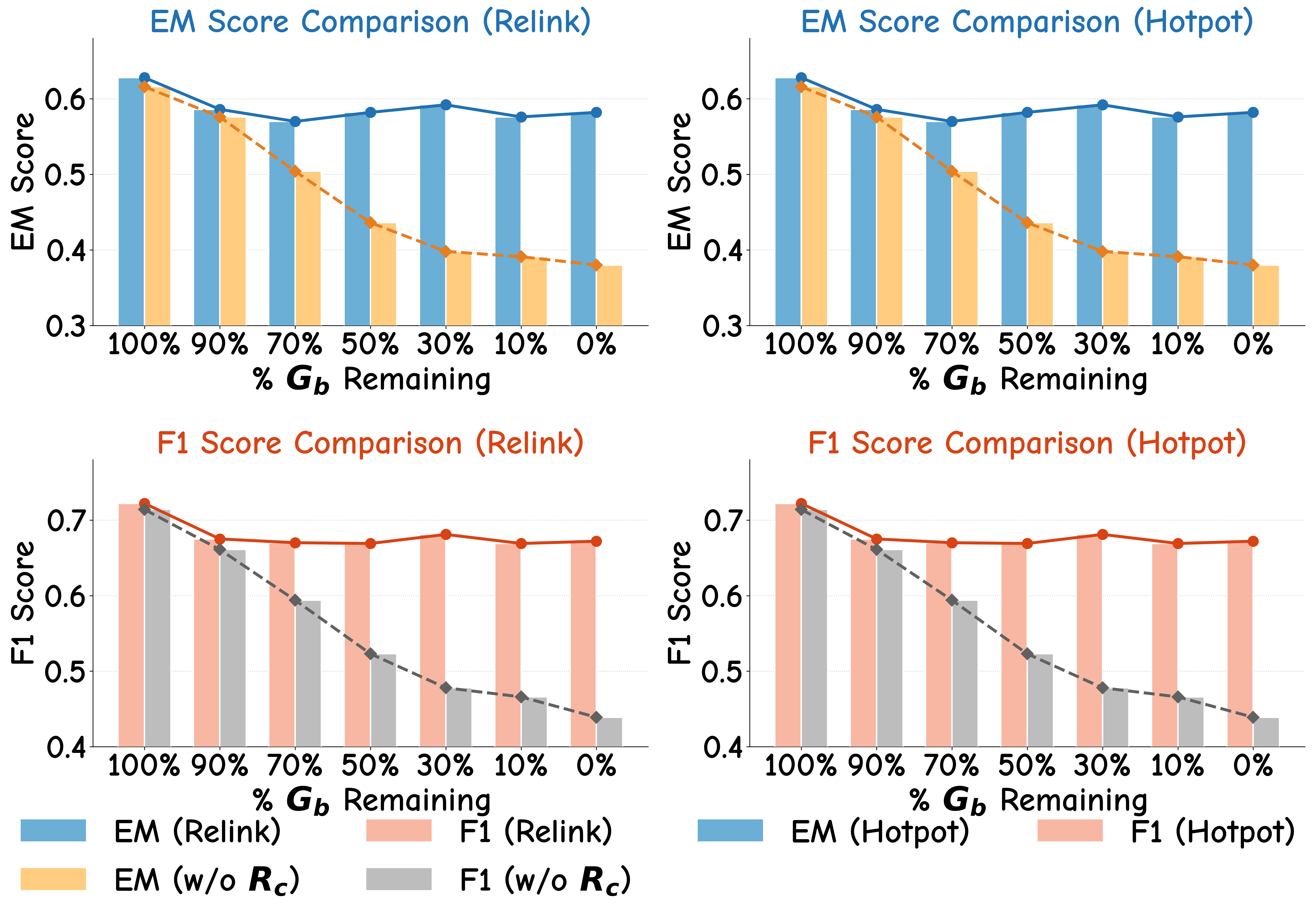}
\caption{Performance trend as the factual graph is reduced. Relink exhibits remarkable robustness to knowledge sparsity, whereas the baseline's performance collapses.}
\label{fig:robustness_trend}
\vspace{-0.5em}
\end{figure}

To simulate increasing knowledge sparsity, we incrementally remove edges from the explicit graph ($\mathcal{G}_b$) and compare the full Relink with the variant without dynamic repair (\textit{w/o} $\mathcal{R}_c$). As shown in ~\Cref{fig:robustness_trend}, Relink consistently outperforms the variant, highlighting the effectiveness of dynamic repair.

\noindent\textbf{The Inherent Brittleness of Static GraphRAG}
Our experiments demonstrate that static reasoning is fundamentally fragile in the presence of incomplete knowledge. The \textit{w/o} $\mathcal{R}_c$ variant, which represents this approach, relies entirely on the integrity of the existing graph structure. As shown in Figure~\ref{fig:robustness_trend}, the F1 score on 2WikiMultiHopQA drops by 34.7\% when 90\% of the edges are removed. This significant decline highlights a key limitation that static reasoning treats paths as static, so the removal of a single crucial edge can cause the entire reasoning process for a query to fail. These results reveal a major weakness in methods that depend solely on static knowledge graphs.

\noindent\textbf{Dynamic Repair Provides Remarkable Resilience.}
In contrast, Relink demonstrates strong adaptability through its \textit{reason-and-construct} paradigm. Even when 90\% of the explicit graph is missing, Relink maintains a high F1 score of 0.669 with only a slight decrease. This robustness is achieved by actively constructing reasoning paths rather than relying on pre-defined routes. Relink leverages the explicit graph as a set of reliable connections, but when these are unavailable, it dynamically explores latent relations from $\mathcal{R}_c$ to bridge the gaps. By flexibly building paths based on available information, Relink overcomes the limitations of static approaches and achieves greater resilience to sparsity in knowledge graphs.

These findings call for a rethinking of robustness in reasoning systems. Instead of assuming a complete knowledge base, robust systems should be built to function effectively with incomplete information. As embodied by the \textit{reason-and-construct} paradigm, dynamic reasoning and adaptive path construction offer a practical and resilient approach for real-world reasoning tasks.

\subsection{Case Study}

To provide a concrete, qualitative illustration of our Relink's mechanism, we analyze its behavior on a specific multi-hop query, addressing RQ4:

\begin{description}
    \item[RQ4] \textbf{\textit{How does Relink’s \textit{reason-and-construct} process operate in practice to overcome knowledge gaps and highly relevant distractor facts where static methods fail?}}
\end{description}

\begin{figure}[htbp]
\centering
\vspace{-0.5em}
\includegraphics[width=1.0\linewidth]{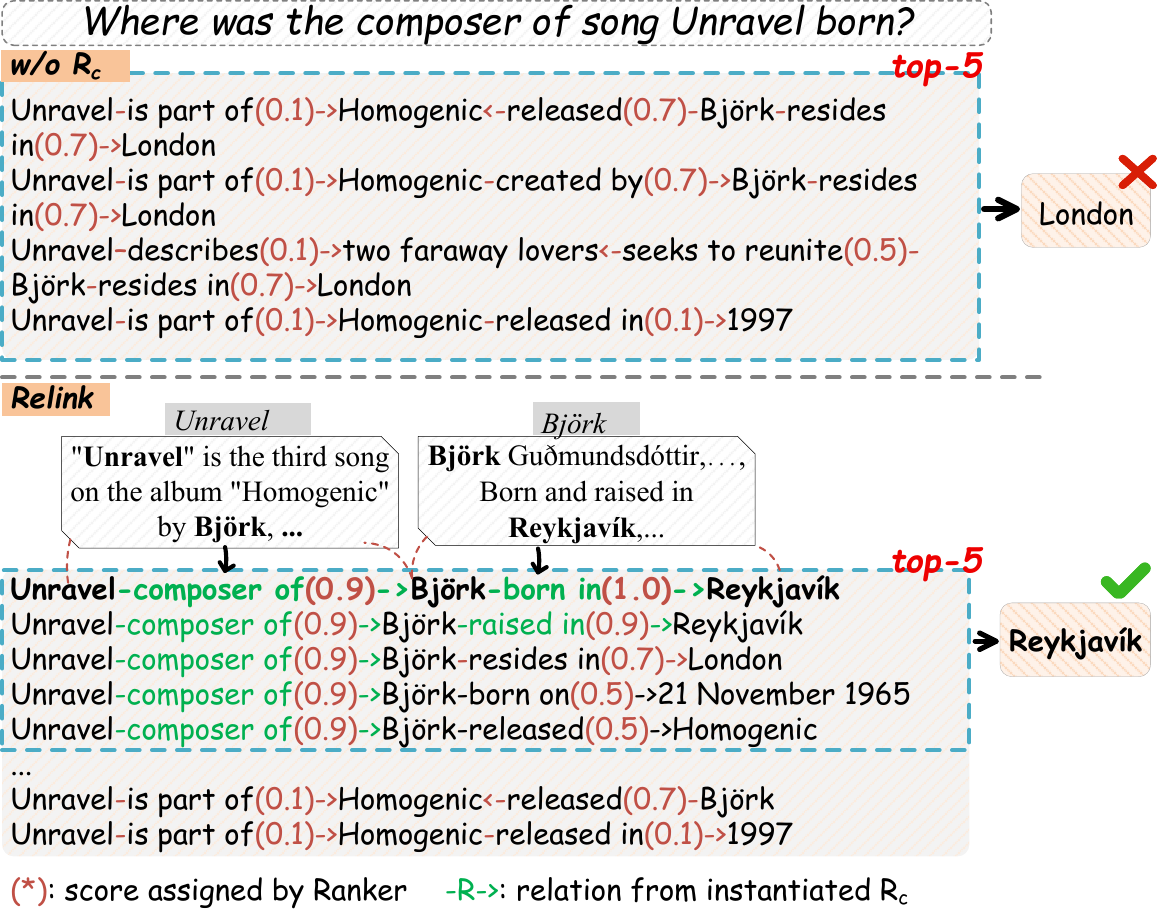}
\caption{A case study contrasting static reasoning with Relink's dynamic approach. The static baseline (\textit{w/o} $\mathcal{R}_c$) is misled by the highly relevant \texttt{resides in} distractor. In contrast, Relink succeeds by dynamically constructing the correct reasoning chain (\texttt{composer of} $\rightarrow$ \texttt{born in}) and using its query-driven ranker to prioritize it.}
\label{fig:case-study}
\vspace{-1.5em}
\end{figure}

The process, depicted in Figure \ref{fig:case-study}, provides a compelling illustration of the fundamental limitations inherent in the dominant \textbf{build-then-reason} paradigm. The baseline model (\textit{w/o} $\mathcal{R}_c$) is fundamentally constrained by what the pre-existing graph \textbf{has}, not what the query \textbf{needs}. Faced with a missing \texttt{composer of} link, it is forced to piece together a low-confidence, circuitous path from available but suboptimal edges. Subsequently, when confronted with multiple facts about the correct entity, it falls into a distractor fact by selecting the highly relevant but incorrect \texttt{resides in} distractor. This failure is a direct consequence of a "one-graph-fits-all" approach, where a static, noisy graph dictates a brittle reasoning process.

In contrast, Relink exemplifies the power of our proposed \textit{reason-and-construct} paradigm, which is guided by what the query \textbf{needs}. It treats the initial KG not as a rigid map, but as a high-precision scaffold to be augmented. Upon identifying the knowledge gap, it proactively uses textual evidence to instantiate the necessary \texttt{composer of} relation. This newly constructed fact is then evaluated in a unified ranking step against existing facts from the KG, including the \texttt{resides in} distractor. The query-driven ranker, seeking to fulfill the specific semantic constraint of “...was \textbf{born}?”, decisively prioritizes the path containing the correct \texttt{born in} relation. This is not mere pathfinding; it is the on-the-fly construction of a compact and query-specific evidence graph.

Ultimately, this comparison highlights a necessary paradigm shift. The case study serves as a microcosm of our central thesis: static reasoning methods are fundamentally limited by their passive, retrieval-based nature, making them vulnerable to both incompleteness and noise. Relink's dynamic approach, which integrates reasoning with on-demand knowledge construction, directly addresses these core challenges. It demonstrates that for robust and precise multi-hop reasoning, the system's focus must shift from navigating what a graph \textbf{has} to intelligently constructing what a query \textbf{needs}.

\vspace{-1.0em}

\section{Conclusion}


In this work, we proposed \textbf{Relink}, a framework designed to challenge the reliance of GraphRAG on static KGs. We proposed a shift to a \textbf{\textit{"reason-and-construct"}} paradigm, where Relink dynamically constructs query-specific reasoning paths by re-linking relations from both explicit KGs and latent text. Our experiments validate this approach: Relink significantly outperforms leading methods and maintains robust performance under the knowledge sparsity that cripples static models. This underscores the value of flexible, on-demand knowledge completion.

\section*{Acknowledgements}
This work was supported by the National Natural Science Foundation of China (Grant No. 62120106008), the Hefei Key Generic Technology Research and Development Program (No. 2024SGJ010), the Youth Talent Support Program of the Anhui Association for Science and Technology (Grant No. RCTJ202420), the Anhui Provincial Science and Technology Fortification Plan (Grant No. 202423k09020015), and the Key Laboratory of Knowledge Engineering with Big Data (the Ministry of Education of China) under Grant No. BigKEOpen2025-03. The computation was completed on the HPC Platform of Hefei University of Technology. Y. He was not supported by any of these funds.

\bibliography{aaai2026}
\makeatletter
\newcommand{\isChecklistMainFile}{}
\makeatother


\end{document}